\documentclass[10pt,twocolumn,letterpaper]{article}

\usepackage[pagenumbers]{cvpr} 

\usepackage{graphicx}
\usepackage{amsmath}
\usepackage{amssymb}
\usepackage{booktabs}

\usepackage{multirow}
\usepackage{booktabs}
\usepackage{adjustbox}
\usepackage{subcaption}
\usepackage{tensor}
\usepackage[utf8]{inputenc}
\usepackage[english]{babel}
\usepackage[ruled,linesnumbered]{algorithm2e}

\usepackage[pagebackref,breaklinks,colorlinks]{hyperref}

\usepackage{amsthm}
\newtheorem{theorem}{Theorem}
\theoremstyle{definition}
\newtheorem{observation}[theorem]{Observation}
\theoremstyle{definition}
\newtheorem{remark}{Remark}

\usepackage[capitalize]{cleveref}
\crefname{section}{Sec.}{Secs.}
\Crefname{section}{Section}{Sections}
\Crefname{table}{Table}{Tables}
\crefname{table}{Tab.}{Tabs.}

\def\cvprPaperID{2619} 
\def\confName{CVPR}
\def\confYear{2022}

\begin{document}

\title{Weakly Supervised Learning of Keypoints for 6D Object Pose Estimation}

\author{Meng Tian\\
National University of Singapore\\
Singapore\\
{\tt\small tianmeng@u.nus.edu}
\and
Gim Hee Lee \\
National University of Singapore\\
Singapore\\
{\tt\small gimhee.lee@nus.edu.sg}
}
\maketitle

\begin{abstract}
State-of-the-art approaches for 6D object pose estimation require large amounts of labeled data to train the deep networks.
However, the acquisition of 6D object pose annotations is tedious and labor-intensive in large quantity.
To alleviate this problem, we propose a weakly supervised 6D object pose estimation approach based on 2D keypoint detection. 
Our method trains only on image pairs with known relative transformations between their viewpoints. Specifically, we assign a set of arbitrarily chosen 3D keypoints to represent each unknown target 3D object and learn a network to detect their 2D projections that comply with the relative camera viewpoints. During inference, 
our network first infers the 2D keypoints from the query image and a given labeled reference image. We then use these 2D keypoints and the arbitrarily chosen 3D keypoints retained from training to infer the 6D object pose.
Extensive experiments demonstrate that our approach achieves comparable performance with state-of-the-art fully supervised approaches.
\end{abstract}

\section{Introduction}
\label{sec:intro}

6D object pose estimation aims to recognize known objects and estimate their 6DoF poses, \ie orientations and translations, in the camera coordinate frame.
It is a crucial component of many applications such as robotic manipulation, augmented reality, and autonomous driving.
Although this problem is easier to solve with RGB-D data \cite{wang2019densefusion, he2020pvn3d}, depth images are not always available.
For example, typical devices for augmented reality (\eg mobile phones and tablets) do not provide depth sensors.
This work focuses on recovering the 6D object pose from single RGB image.

Recent deep learning-based approaches either directly regress the 6D pose from an image \cite{xiang2018posecnn, li2019cdpn} or build 2D-3D correspondences and indirectly solve for the pose via PnP \cite{peng2019pvnet, zakharov2019dpod, park2019pix2pose}.
Although achieving good performances, these works rely on full 6D pose annotations and well-textured 3D CAD models.
However, labeling 6D poses at large scale is a tedious, costly and labor-intensive chore.
Another category of works \cite{sundermeyer2018implicit, zakharov2019dpod, park2020neural} makes use of the available object model to reduce the labeling effort.
They render images with the CAD model and train the pose estimator on the synthetic images.
Unfortunately, this introduces a domain shift problem between real and synthetic images that is non-trivial to resolve.
As a result, the performances of these methods are not ideal.

\begin{figure}[t]
\begin{center}
\includegraphics[width=\columnwidth]{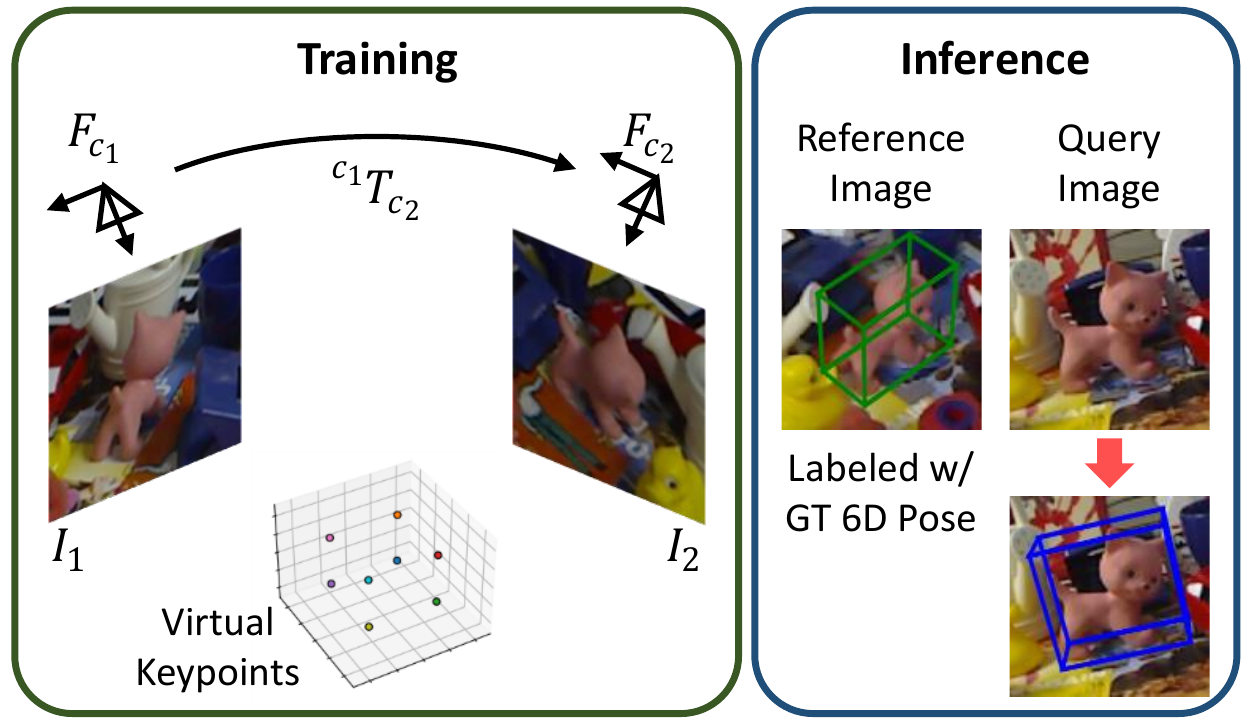}
\end{center}
\vspace{-3mm}
\caption{Our weakly supervised 6D object pose estimation setting. The relative transformation $\tensor[^{c_1}]{T}{_{c_2}}$ and randomly generated 3D virtual keypoints are used for training. We then infer the 6D object pose of a query image using a given labeled    
reference image and 3D virtual keypoints. 
}
\label{fig:teaser}
\vspace{-5mm}
\end{figure}

In \cite{cai2020reconstruct, zhao2020learning}, multi-view geometry is used as an alternative supervisory signal. 
Given a pair of images that observe the same object, the 6D object poses in each image must comply to the relative transformation between the two image viewpoints.
Relative transformations between camera viewpoints are relatively easier to obtain.
For example, image pairs can be captured by a binocular camera where the relative transformation is given by camera calibration.
In the scenario of robotic manipulation, the relative transformations can be computed from forward kinematics.
RLLG \cite{cai2020reconstruct} establishes 2D-3D correspondences and estimates the 6D pose of the object using the perspective-n-point (PnP) algorithm.
The relative transformation is used to ensure that the predicted 3D object coordinates for the same object point are consistent across different views.
OK-POSE \cite{zhao2020learning} automatically discovers a set of 3D keypoints which satisfies the relative transformation.
The 6D pose is inferred from the corresponding keypoints in the query and reference image.
However, the former still requires full 6D object poses during training and the latter can predict 3D keypoints that are inconsistent across different views.

In this paper, we develop a weakly supervised 6D object pose estimation method based on 2D keypoint detection.
In contrast to previous approaches that train on ground truth 6D object poses, our method learns from the relative transformations between the viewpoints of image pairs as illustrated in Fig.~\ref{fig:teaser}.
Since we do not have the ground truth 3D keypoints during training, we assign a set of randomly generated 3D points which we refer to as the ``virtual keypoints'' to represent the object.
The problem is then reduced to train a network to detect the corresponding 2D projections of the virtual keypoints in the images.
Our network is trained by minimizing the errors between the triangulated keypoints and the virtual keypoints up to a rigid transformation.
We do the 3D triangulation with the 2D keypoints predicted by the network and the ground truth relative transformation. 

It is worthy to note that there exists an offset rigid transformation between the coordinate frame of randomly generated virtual keypoints and the coordinate frame of the 3D object.
Fortunately, this offset rigid transformation remains fixed regardless of the image viewpoint, and it can be recovered from 
a given reference image with the 6D object pose.
Our reference image follows the setting of \cite{zhao2020learning}, where the 6D object pose can be easily obtained from simple epipolar geometry 
(details in the supplementary). Furthermore, only one reference image is needed for the entire inference stage. 
During inference, we first infer the 2D keypoints of the reference and query images.
We then compute the offset rigid transformation using the virtual keypoints retained from training, the 2D keypoints and the 6D object pose of the reference image.
Finally, we compute the 6D object pose in the query image with the pose obtained from PnP on the virtual and query image 2D keypoints compensated with the offset rigid transformation.

We evaluate our approach on two standard benchmarks: LINEMOD \cite{hinterstoisser2012model} and OCCLUSION \cite{brachmann2014learning}.
Our approach achieves significantly better performance than OK-POSE~\cite{zhao2020learning}.
We are also comparable to the state-of-the-art methods that are trained on real images annotated with full 6D poses. Our main contributions in this work are:
\vspace{-1mm}
\begin{itemize}
\item We propose a novel approach based on 2D keypoint detection for 6D object pose estimation. Our method uses only relative transformations of image pairs for weak supervision.
\vspace{-1mm}
\item Our method  significantly outperforms the existing work on a similar weakly supervised  setting~\cite{zhao2020learning}, and is comparable with state-of-the-art fully supervised approaches trained on real images annotated with ground truth 6D object poses.
\end{itemize}

\begin{figure*}[t]
\begin{center}
\includegraphics[width=\linewidth]{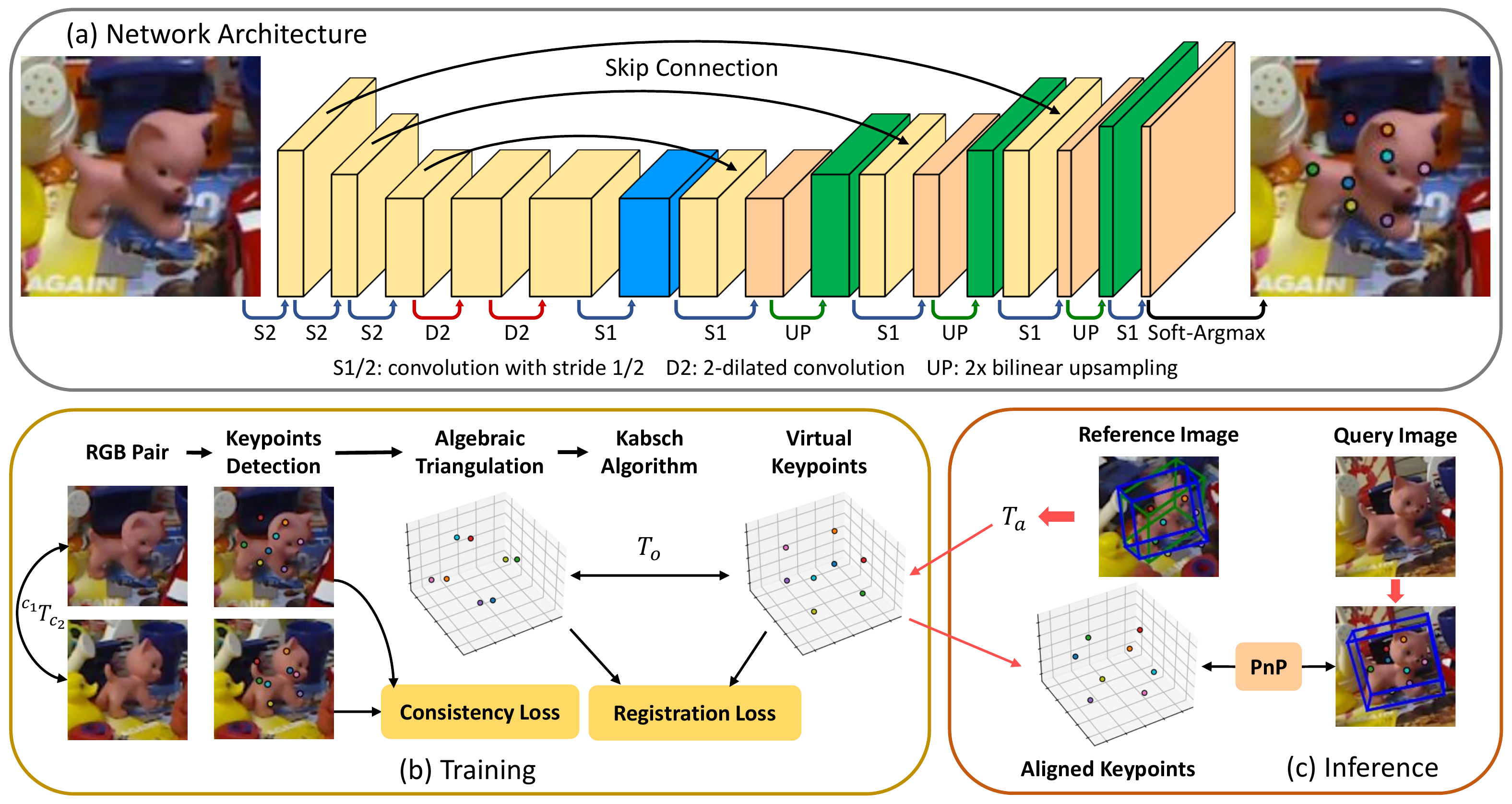}
\vspace{-6mm}
\end{center}
\caption{(a) Architecture of our 2D keypoint detection network (\S~\ref{sect:network}). (b) The 3D virtual keypoints (\S~\ref{sect:VirtualKpts}), and registration and consistency losses (\S~\ref{sect:learning}) are proposed for training. (c) A reference image labeled with 6D object pose and the 3D virtual keypoints are used to estimate the 6D object pose in a query image during inference (\S~\ref{sect:alignment}).
}
\label{fig:network}
\vspace{-3mm}
\end{figure*}

\section{Related Work}
Due to its practical importance, there is a large body of works on 6D object pose estimation.
We first review recent works in monocular 6D pose estimation and then we discuss several important works on reducing the labeling effort by using synthetic data or weak labels.

\vspace{-4mm}
\paragraph{Fully Supervised Approaches.}
Recent deep learning-based pose estimation approaches can be roughly divided into three categories.
The first category establishes 2D-3D correspondences between the input image and the 3D object model, and computes the 6D pose via solving a PnP problem.
\cite{rad2017bb8, tekin2018real, oberweger2018making, hu2018segmentation} detect the corners of the tight 3D bounding box in images.
PVNet \cite{peng2019pvnet} and its extensions \cite{song2020hybridpose, he2020pvn3d} predict the 2D projections of a sparse point set selected on the surface of the object model.
\cite{li2019cdpn, park2019pix2pose, zakharov2019dpod} densely map each foreground pixel to its corresponding point in the object model.
The second category directly regresses 6D pose from the input image.
Both \cite{kehl2017ssd} and \cite{li2018unified} discretize the pose space and turn pose estimation into a classification problem.
Additionally, \cite{li2018unified} employs residual regression to achieve fine pose estimation.
\cite{xiang2018posecnn, wang2019densefusion} propose a point matching loss to deal with ambiguities induced by symmetric objects, while \cite{manhardt2019explaining} handles ambiguities by generating multiple pose hypotheses.
The third category retrieves the closest viewpoint from a codebook based on the input image.
\cite{kehl2016deep} trains an autoencoder to extract a feature for each image patch, which is further used to cast vote in 6D pose space.
\cite{sundermeyer2018implicit} uses an augmented autoencoder to learn a latent representation for the rotation. Despite achieving superior performances, these methods require large amounts of real annotated data to train the deep networks.
Our method also relies on keypoint detection, but we only requires weak labels to train the network.

\vspace{-4mm}
\paragraph{Domain Adaptation.}
Training deep networks on synthetic data is a promising solution to reduce the labeling effort. However, the domain shift between real and synthetic data causes drastic drops in performance.
Several works are proposed to bridge the domain gap for 6D pose estimation.
These methods use unlabeled real images to improve the performance of labeled synthetic images.
PixelDA \cite{bousmalis2017unsupervised} uses a conditional GAN to adapt the synthetic training images such that they appear like real images.
SynDA \cite{planche2019seeing} learns to map unseen real images into the geometric domain (\eg normal maps) in which the pose estimator is trained.
\cite{rad2018domain} and \cite{rad2018feature} train a feature mapping network to map the latent features of real images into the synthetic feature space before using them as inputs to the pose estimator.
Self6D \cite{wang2020self6d} is first trained in a fully supervised manner on synthetic RGB images.
It is then fine-tuned on unlabeled real RGB-D data in a self-supervised manner.
DeceptionNet \cite{zakharov2019deceptionnet} tackles the domain adaptation using domain randomization techniques.
It consists of multiple differentiable perturbation modules, which alters the input images and forces the pose estimation network to be robust to domain changes.
Despite the progresses made by these approaches, bridging the domain gap remains a difficult challenge.
Consequently, their performances remain unsatisfactory compared to approaches trained on real data.
Although our approach is trained on real images, we reduce labeling efforts by using labels that are much easier to obtain.

\vspace{-4mm}
\paragraph{Weakly Supervised Approaches.}
The most recent works focus on learning 6D object pose estimation with weak labels such as the relative transformation between the viewpoints of an image pair.
RLLG \cite{cai2020reconstruct} predicts corresponding 3D object coordinates for foreground pixels of the target object without assuming any prior knowledge of the CAD model.
The established dense 2D-3D correspondences can be viewed as a partial shape reconstruction in object coordinate frame.
Additional relative transformations are used to ensure that the reconstructed shapes are consistent across different views.
OK-POSE \cite{zhao2020learning} further relaxes the requirement of 6D pose labels.
It automatically discovers a set of 3D keypoints that are visually consistent across different views and geometrically consistent with the relative transformations.
The 6D pose is then inferred from the corresponding keypoints in a query image and a reference image. Our method is inspired by OK-POSE.
However, our network learns to detect the 2D projections of arbitrarily chosen 3D points which are associated with the object coordinate frame after training via a single annotated image.

\section{Our Method}
Given an RGB image, the task of 6D object pose estimation is to detect each object, and estimate its location and orientation in the 3D space.
We adopt the correspondence-based approach, which detects the 2D image projections of 3D keypoints and computes the 6D pose via the PnP algorithm.
As illustrated in Fig.~\ref{fig:network}, our weakly supervised approach consists of two stages: 1) An object detector, \eg YOLOv3 \cite{redmon2018yolov3} is used to estimate the location and size of each object in the image. 2) The region of interest is cropped according to its size and location, and fed to a convolutional network (\S~\ref{sect:network}) to detect the 2D keypoints.
Our keypoint detection network is trained on images pairs with known relative transformations between their viewpoints and randomly generated 3D keypoints (\S~\ref{sect:VirtualKpts} \& \ref{sect:learning}).
During inference, the 6D object pose in a query image is computed from the detected keypoints and a reference image 
(\S~\ref{sect:alignment}).

\subsection{2D Keypoint Detection Network}
\label{sect:network}

As shown in Fig.~\ref{fig:network}(a), the backbone of our keypoint detection network is a ResNet-18 \cite{he2016deep} followed by a series of up-sampling and $3 \times 3$ convolution layers.
We also make same modifications as PVNet \cite{peng2019pvnet}. Specifically, after the feature map is down-sampled to $1/8$ of the original size, we discard the subsequent pooling layers and replace subsequent convolution layers with dilated ones to maintain the receptive fields.
Skip connections are applied between features maps of same resolutions in the down-sampling and up-sampling stages.
The network has $N$ output channels, with each channel being a heatmap $p_i(h, w)$ which represents the probability of the $i$-th keypoint being at the pixel $(h, w)$.
The 2D keypoint location is then obtained by taking the soft-argmax \cite{sun2018integral} operation:
\begin{equation}
  (u_i, v_i) = \sum_{h, w} \big(h \cdot p_i(h, w), w \cdot p_i(h, w)\big).
  \label{eq:keypoints}
\end{equation}

\subsection{3D Virtual Keypoints}
\label{sect:VirtualKpts}
\begin{figure}[t]
\begin{center}
\includegraphics[width=0.7\columnwidth]{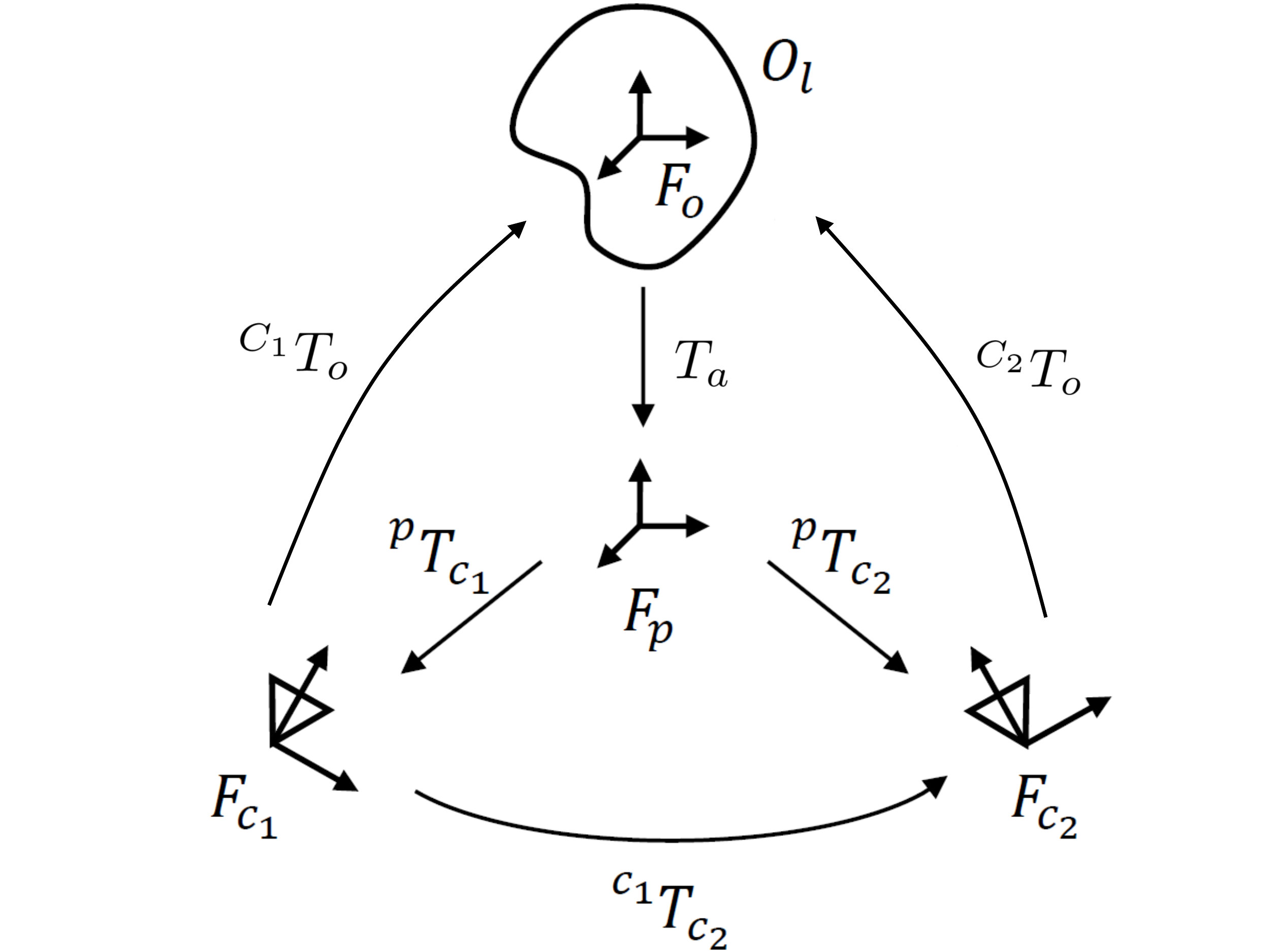}
\vspace{-3mm}
\end{center}
\caption{Visualization of the relative transformations.}
\label{fig:relative_pose}
\vspace{-3mm}
\end{figure}

We assume that the training data is image pairs $(I_1, I_2)$, where each image pair observes the same object $O_l$.
The relative transformation $^{c_1}T_{c_2} \in \operatorname{SE}(3)$ between the two viewpoints is known.
Each image is additionally annotated with the object class label $l$ and 2D bounding box.
We propose the use of randomly generated 3D ``virtual keypoints" as surrogates due to the absence of the ground truth 3D keypoints. 
Let us denote the coordinate frame of these 3D virtual keypoints as $F_p$. We generate a set of 3D virtual keypoints for each object class. The generated 3D virtual keypoints are used in training and kept for inference. Using training data that consists of only pairs of images $(I_1, I_2)$ and their associated relative transformations $^{c_1}T_{c_2}$, we then train our 2D keypoint detection network to detect 2D keypoints in the images $I_1$ and $I_2$ that form 2D-3D correspondences with the 3D virtual keypoints, respectively. As illustrated in Fig.~\ref{fig:relative_pose}, the 6D poses of the respective images to the coordinate frame of the 3D virtual keypoints, \ie $^{p}T_{C_1}$ and $^{p}T_{C_2}$ can be computed from PnP using the 2D-3D correspondences.

\begin{observation}
\label{obs:offsetTransformation}
Since the 3D virtual keypoints are randomly generated without any knowledge of the 3D object, there exists an offset rigid transformation $T_a$ from the coordinate frame of the 3D virtual keypoints $F_p$ to the coordinate frame of the 3D object $F_O$. This offset rigid transformation $T_a$ is independent of the images and stays fixed. 
\end{observation}

\begin{remark}
\label{remark:offsetTransformation}
Referring to Fig.~\ref{fig:relative_pose}, we are given a reference image $F_{C_1}$, the 6D object pose $^{C_1}T_{o}$ in $F_{C_1}$, and a query image $F_{C_2}$ during inference. The 6D poses $^{p}T_{C_1}$ and $^{p}T_{C_2}$ of $F_{C_1}$ and $F_{C_2}$ can be computed with the 2D keypoints from our network and the 3D virtual keypoints using PnP. The offset rigid transformation can then be computed as: $T_a={^{C_1}T_{o}^{-1}}^{p}T_{C_1}^{-1}$.
Finally, the required 6D object pose in the query image $F_{C_2}$ is given by: $^{C_2}T_{o}={^{p}T_{C_2}^{-1}}T_a^{-1}$.
\end{remark}

\begin{remark}\label{remark:oneRefImage}
Since $T_a$ is independent of the images and stays fixed, only one reference image $F_{C_1}$ with 6D object pose $^{C_1}T_o$ is needed for the entire inference stage. 
\end{remark}

\subsection{Weak Supervision from Two-View Geometry}
\label{sect:learning}
Given training data that consists of image pairs with their associated relative transformations and the randomly generated 3D virtual keypoints, 
the objective is now reduced to training a network to detect 2D keypoints from each image pair that: 1) the triangulated 3D points align well with the 3D virtual keypoints, and 2) satisfy the two-view epipolar constraint.
To this end, we propose two losses: a registration loss and a consistency loss as shown in Fig.~\ref{fig:network}(b).

\vspace{-4mm}
\paragraph{Registration Loss.}
We design the registration loss for the network to learn to detect 2D keypoints that correspond to the pre-defined 3D virtual keypoints.
To this end, we propose to first estimate the 3D positions of the keypoints from their 2D estimates using triangulation. 
Subsequently, the error between the triangulated 3D keypoints and the 3D virtual keypoints are minimized to enforce the cross-view consistency between the 2D keypoints detected on two images. We use differentiable triangulation to ensure that our registration loss can be used to train our network.

We denote the camera intrinsic matrices of the image pair $(I_1, I_2)$ as $K_1$ and $K_2$, respectively.
The 2D keypoints on $I_1$ and $I_2$ are denoted as $(u^1_i, v^1_i)$ and $(u^2_i, v^2_i)$, where $i \in [1, \dots , N]$ is the index of the 2D keypoints.
Each 2D keypoint is treated independently in the triangulation algorithm.
$\hat{\mathbf{x}}_i$ represents the $i$-th triangulated 3D keypoint, which corresponds to virtual keypoint $\mathbf{x}_i$.
Both $\hat{\mathbf{x}}_i$ and $\mathbf{x}_i$ are homogeneous coordinates with unit fourth coordinate.
We use a linear algebraic triangulation approach \cite{iskakov2019learnable} to compute $\hat{\mathbf{x}}_i$, \ie
    $A_i \hat{\mathbf{x}}_i = \mathbf{0}$,
where $A_i$ is a matrix computed from the full projection matrices $(P_1, P_2)$ and the 2D keypoint detections $(u^1_i, v^1_i)$, $(u^2_i, v^2_i)$.
We assume that $\hat{\mathbf{x}}_i$ is in the camera coordinate frame of $I_1$.
The full projection matrices $P_1$ and $P_2$ are:
\begin{equation}
    P_1 = K_1 [I \mid \mathbf{0}], \quad P_2 = K_2 [R \mid \mathbf{t}], 
\end{equation}
where $R= {^{C_1}R_{C_2}^{\top}}$ and $\mathbf{t}= -{^{C_1}R_{C_2}^{\top}} {^{C_1}\mathbf{t}_{C_2}}$ are computed from the rotation matrix and translation vector from the relative transformation label $^{C_1}T_{C_2}$ between $(I_1,I_2)$. As a result, $A_i$ is computed as:
\begin{equation}
    \small A_i = \big[
    [u^1_i \mathbf{p}^{3}_1 - \mathbf{p}^{1}_1]^\top, [v^1_i \mathbf{p}^{3}_1 - \mathbf{p}^{2}_1]^\top, [u^2_i \mathbf{p}^{3}_2 - \mathbf{p}^{1}_2]^\top, [v^2_i \mathbf{p}^{3}_2 - \mathbf{p}^{2}_2]^\top
    \big]^\top,
\end{equation}
where $\mathbf{p}_l^k$ is the $k$-th row of projection matrix $P_l$.
$\hat{\mathbf{x}}_i$ is solved through differentiable Singular
Value Decomposition: $A_i = U \Sigma V^{\top}$.
It is set to the last column of $V$ and further divided by its fourth coordinate.

We use the Kabsch algorithm \cite{kabsch1976solution} to estimate an optimal transformation $T_o \in \operatorname{SE}(3)$ that aligns the two sets of correspondent 3D keypoints $\{\hat{\mathbf{x}}_i \leftrightarrow \mathbf{x}_i\}$ that are in different coordinate frames.
The registration loss is then formulated as the minimization of the error between the triangulated 3D keypoints $\hat{\mathbf{x}}_i$ and the correspondent 3D virtual keypoints $\mathbf{x}_i$ aligned by the rigid transformation $T_o$, i.e:
\begin{equation}
\mathcal{L}_{reg} = \frac{1}{N} \sum_{i} \| \hat{\mathbf{x}}_i - T_o \mathbf{x}_i \| _ 2 .\vspace{-5mm}
\end{equation}
\vspace{-4mm}
\paragraph{Consistency Loss.}
We propose a consistency loss to enforce the two-view epipolar constraint.
This loss is shown to improve the accuracy of keypoint localization.
The 2D keypoints $(u^1_i, v^1_i)$ and $(u^2_i, v^2_i)$ correspondence from two views must satisfy the epipolar constraint \cite{hartley2003multiple}. 
Consequently, we define the consistency loss as:
\begin{equation}
 \mathcal{L}_{con} = \frac{1}{N} \sum_{i}[u_i^2, v_i^2, 1] F [u_i^1, v_i^1, 1]^{\top}, \vspace{-2mm}
\end{equation}
where $F = K_2^{-\top} [\mathbf{t}]_{\times} R K_1^{-1}$ is the fundamental matrix computed from the intrinsic matrices and $\tensor[^{c_1}]{T}{_{c_2}}$.

Finally, the overall objective used to train the keypoint detection network is given by:
    $\mathcal{L} = {\lambda}_1 \mathcal{L}_{reg} + {\lambda}_2 \mathcal{L}_{con}$,
where $\lambda_1$ and $\lambda_2$ are used to weigh the respective loss terms.

\subsection{Inference}
\label{sect:alignment}
As illustrated in Fig.~\ref{fig:network}(c), we are given a reference image 
with the 6D object pose (green box) and the 3D virtual keypoints (retained from training) during inference. Refer to our supplementary for details on how to get the reference image. Given a query image, the goal is to find the 6D object pose in the query image camera frame. 

We first apply an object detector on the reference and query images, and then pass the cropped image according to the object detection bounding boxes into our network for 2D keypoint detection. Subsequently, we use PnP to compute the 6D pose of the 3D virtual keypoints in the reference and query image frames, respectively. The 6D poses of the 3D virtual keypoints are illustrated by the blue boxes in Fig.~\ref{fig:network}(c).
According to Observation~\ref{obs:offsetTransformation}, there exists a fixed offset transformation $T_a$ (\cf Fig.~\ref{fig:relative_pose}) from the reference camera to the 3D object. $T_a$ is illustrated by the discrepancy between the green and blue boxes in the reference image in Fig.~\ref{fig:network}(c). Finally, we compute the 6D object pose in the query image by compensating for the offset transformation $T_a$ with the 6D pose of the 3D virtual keypoints to the query image. 
Refer to Remark~\ref{remark:offsetTransformation} for the full details.

We compute the average offset transformation $\bar{T}_a$ for higher accuracy
when more than one reference image with 6D pose labels are available during inference.
Since the predicted poses are always noisy, we implement the RANSAC algorithm \cite{fischler1981random} to select inliers from $\{R^i_a, \mathbf{t}^i_a\}$.
Inliers are defined to be those transformations whose rotation distance and translation distance to the hypothesis are lower than the thresholds.
The average offset rotation $\bar{R}_a$ and translation $\bar{\mathbf{t}}_a$ are obtained by taking the geodesic $L_2\text{-Mean}$ \cite{hartley2013rotation} and the arithmetic mean, respectively.
The algorithm is summarized in the supplementary material.

\section{Experiments}

\paragraph{Datasets.}
LINEMOD \cite{hinterstoisser2012model} contains 13 texture-less objects placed in cluttered scenes.
This dataset is the \textit{de facto} 
benchmark for 6D pose estimation of single object without occlusion.
Following \cite{brachmann2016uncertainty, rad2017bb8, wang2019densefusion}, $\sim$ 15\% of the images which covers different viewpoints of each object are selected for training.
The remaining images are used for evaluation.
OCCLUSION \cite{brachmann2014learning} is a sequence of LINEMOD, in which all visible objects are annotated with ground truth poses.
This dataset is suitable for evaluating pose estimators when objects are under occlusion.
YCB-Can is a subset of the YCB-Video dataset \cite{xiang2018posecnn}.
It consists of images containing the ``010-potted-meat-can" object.
In total, there are 16,992 training images, which we generate 8,496 image pairs without reusing the images in different pairs.
For evaluation, we select 766 keyframes from the original test sequences.

\vspace{-4mm}
\paragraph{Evaluation Metrics.}
We use the standard metric ADD(-S) \cite{hinterstoisser2012model} to evaluate the quality of the pose estimation.
The ADD metric is calculated as the average distance between model vertices under the estimated pose $(R_p, \mathbf{t}_p)$ and the ground truth pose $(R_{\text{gt}}, \mathbf{t}_{\text{gt}})$:
\begin{equation}
\operatorname{ADD} = \underset{\mathbf{x} \in \mathcal{M}}{\text{avg}} \| (R_p \mathbf{x} + \mathbf{t}_{p}) - (R_{\text{gt}} \mathbf{x} + \mathbf{t}_{\text{gt}}) \|_2 \,, \vspace{-2mm}
\end{equation}
where $\mathcal{M}$ represents the 3D model of the object.
To handle symmetric objects, the ADD metric is extended to the distance between closest model points:
\begin{equation}
 \operatorname{ADD-S} = \underset{\mathbf{x}_1 \in \mathcal{M}}{\text{avg}} \min_{\mathbf{x}_2 \in \mathcal{M}} \| (R_p \mathbf{x}_1 + \mathbf{t}_p) - (R_{\text{gt}} \mathbf{x}_2 + \mathbf{t}_{\text{gt}}) \|_2 \,. 
\end{equation}
Conventionally, a pose is considered to be correct if its ADD metric is less than $10\%$ of object diameter.

\vspace{-4mm}
\paragraph{Implementation Details.}
The width $w$, height $h$, and location $(c_x, c_y)$ of the 2D bounding box are used to crop the region of interest for 2D keypoint localization.
To keep the aspect ratio, the size of the region is set to $1.5 \cdot \max (h, w)$.
The cropped RGB patch is further resized to $192 \times 192$ and fed to the keypoint detection network.
We set the hyperparameters of the overall loss to $\lambda_1 = 100$ and $\lambda_2 = 100$.
On LINEMOD and OCCLUSION, we use exactly the same training and test splits as prior works \cite{rad2017bb8, tekin2018real, peng2019pvnet}.
As a result, there are only around 180 training images per class.
To prevent the network from overfitting to the background and simulate occlusions, we segment the foreground objects in the training images and paste them on random images from the SUN Database \cite{xiao2010sun}.
During training, we add random noise sampled from truncated normal distribution to the location and size of the ground truth bounding box to account for the errors of the 2D object detector.
Image pairs are randomly generated from  the training images.
Although our approach does not require 6D object pose annotations for training, the relative transformation between the viewpoints of image pairs are computed from ground truth 6D poses since they are not provided in the original dataset.
An improved version of YOLOv3 implemented by the Ultralytics \cite{glenn_jocher_2020_4154370} is employed as our 2D object detector.
The training data is generated using the same copy-and-paste technique.
In practice, we find that the detector easily overfits to foreground objects.
Additionally, we randomly select $10,000$ images from the photo-realistic synthetic dataset \cite{hodavn2020bop} and add them to the training set.
Note that the synthetic data is only used to train the object detector.
\begin{figure*}[t]
\centering
  \begin{subfigure}{0.48\linewidth}
    \centering
    \includegraphics[width=\columnwidth]{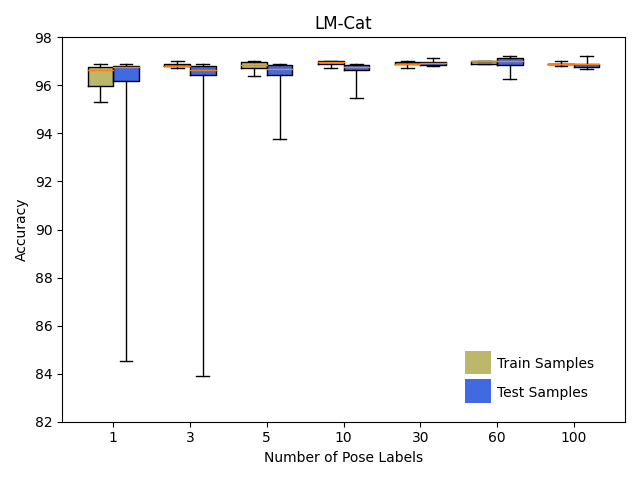}
  \end{subfigure}
  \begin{subfigure}{0.48\linewidth}
    \centering
    \includegraphics[width=\columnwidth]{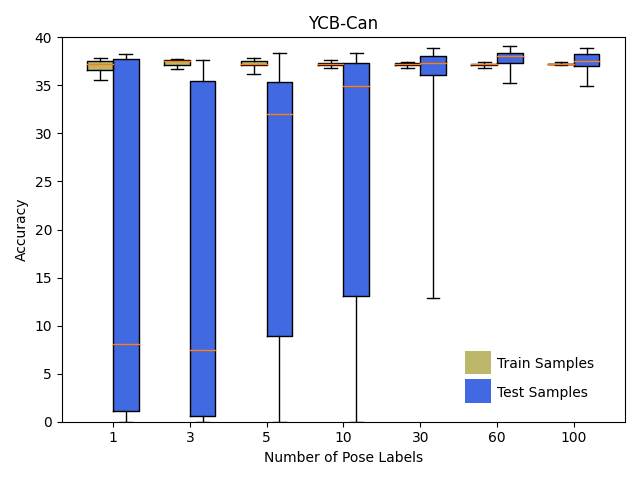}
  \end{subfigure}
  \vspace{-5mm}
\caption{Accuracy w.r.t. the number of pose labels. The box corresponds to the second and third quartile. The solid line in the box depicts the median. The whiskers mark the minimum and maximum accuracies, respectively.}
\label{fig:boxplot}
\end{figure*}

\subsection{Ablation Studies}

Ablation studies are conducted on the ``cat" sequence to evaluate the contribution of each component in our method.

\vspace{1mm}
\noindent\textbf{Fused Training Data.}
LINEMOD 
has limited number of training images.
We use additional training images generated by copy-and-paste technique (\cf ``Fuse" in Tab.~\ref{table:ablation_results}).
To validate the necessity, we train the keypoint detection network with and without using ``Fuse".
Although our pose estimator can achieve highest accuracy on LINEMOD when trained without ``Fuse", it completely fails on OCCLUSION.
Considering that OCCLUSION contains test images which are free from occlusion, training with ``Fuse" data enables our pose estimator to handle object occlusions, and also prevents it from overfitting to background.

\vspace{1mm}
\noindent\textbf{Consistency Loss.}
As indicated in Tab.~\ref{table:ablation_results}, the registration loss itself is capable of training the keypoint detection network.
Together with the consistency loss which regularizes the 2D keypoints correspondences between two views via fundamental matrix, our pose estimator performs significantly better, especially when objects are under occlusion.

\vspace{1mm}
\noindent\textbf{Number of Keypoints.}
To evaluate the influence of the number of keypoints on our pose estimator, we train the network with 4, 6, 8, and 12 keypoints, respectively.
Since the 3D object model is not available, we define the 3D keypoints as the vertices of regular polyhedron (\eg tetrahedron, octahedron, cube, and icosahedron) such that the keypoints are evenly distributed in 3D space.
The results in Tab.~\ref{table:ablation_results} show that the accuracy of our pose estimator increases with the the number of keypoints and peaks at 8.
In contrast to previous fully supervised method \cite{peng2019pvnet} with increasing accuracy after 8 keypoints, our accuracy drops when there are too many keypoints, \ie 12.
We conjecture two reasons for this observation: 1) weak supervision makes it harder for the network to learn to detect more keypoints; 2) unlike \cite{peng2019pvnet} where the keypoints are obtained from the 3D object, our 3D virtual keypoints are randomly generated and thus have weak relation to the 3D object.

\vspace{1mm}
\noindent\textbf{Distribution of Keypoints.}
We choose evenly spaced virtual keypoints since there is no ground truth 3D keypoints.
However, the regular distribution of virtual keypoints is not necessary for our method to achieve good performance.
In addition to the vertices of regular polyhedron, we test our method with keypoints randomly sampled on sphere (``r-sphere” in Tab.~\ref{table:ablation_results}) and within a cubic volume (``r-volume”).
The accuracies do not differ too much on the two datasets. 

\begin{table}[t]
\small
\begin{center}
\begin{adjustbox}{max width=\columnwidth}
\begin{tabular}{ c | c | c  c | c  c}
\toprule
\multirow{2}{*}{Fuse} & \multirow{2}{*}{$L_{con}$} & \multicolumn{2}{c|}{Virtual Keypoints} & \multicolumn{2}{c}{Accuracy} \\
\cline{3-6}
 &  & num. & distrib. & {LM} & {OCC} \\
\midrule
 & \checkmark & 8 & regular & \textbf{97.31} & 0.0 \\
\checkmark & & 8 & regular &93.71 & 8.23 \\
\checkmark & \checkmark & 8 & regular & 97.01 & \textbf{25.17} \\
\checkmark & \checkmark & 8 & r-sphere & 96.11 & 24.38 \\
\checkmark & \checkmark & 8 & r-volume & 97.21 & 22.20 \\
\checkmark & \checkmark & 4 & regular & 38.32 & 5.25 \\
\checkmark & \checkmark & 6 & regular & 96.21 & 17.84 \\
\checkmark & \checkmark & 12 & regular & 96.41 & 16.15 \\
\bottomrule
\end{tabular}
\end{adjustbox}
\vspace{-3mm}
\end{center}
\caption{
  Ablations on ``cat" sequence. We report the accuracies in terms of ADD(-S) metric on both LINEMOD and OCCLUSION.}
\vspace{-6mm}
\label{table:ablation_results}
\end{table}

\begin{table*}[t]
\small
\begin{center}
\begin{adjustbox}{max width=\linewidth}
\begin{tabular}{ c | c  c | c  c  c | c  c  c}
\toprule
\multirow{2}{*}{methods} & \multicolumn{2}{c|}{Relative Pose} & \multicolumn{3}{c|}{Synthetic Data} & \multicolumn{3}{c}{RGB with Pose Annotations} \\
\cline{2-9}
& {OK-POSE \cite{zhao2020learning}} & \textbf{Ours} & {AAE \cite{sundermeyer2018implicit}} & {DPOD \cite{zakharov2019dpod}} & {NOL \cite{park2020neural}} & {PVNet \cite{peng2019pvnet}} & {CDPN \cite{li2019cdpn}} & {DPOD \cite{zakharov2019dpod}} \\
\midrule
ape & 35.8 & \textbf{89.14} & 3.96 & 37.22 & 35.4 & 43.62 & 64.38 & 87.73 \\
benchvise & 26.1 & 99.61 & 20.92 & 66.76 & 55.6 & \textbf{99.90} & 97.77 & 98.45 \\
cam & 34.7 & \textbf{98.14} & 30.47 & 24.22 & 37.5 & 86.86 & 91.67 & 96.07 \\
can & 22.6 & 99.02 & 35.87 & 52.57 & 65.5 & 95.47 & 95.87 & \textbf{99.71} \\
cat & 32.2 & \textbf{97.01} & 17.90 & 32.36 & 38.1 & 79.34 & 83.83 & 94.71 \\
driller & 28.5 & \textbf{99.01} & 23.99 & 66.60 & 52.2 & 96.43 & 96.23 & 98.80 \\
duck & 28.5 & \textbf{87.14} & 4.86 & 26.12 & 14.7 & 52.58 & 66.76 & 86.29 \\
eggbox* & 41.3 & \textbf{100.0} & 81.01 & 73.35 & 93.7 & 99.15 & 99.72 & 99.91 \\
glue* & 32.2 & 99.42 & 45.49 & 74.96 & 63.1 & 95.66 & \textbf{99.61} & 96.82 \\
holepuncher & 15.0 & \textbf{92.67} & 17.60 & 24.50 & 34.4 & 81.92 & 85.82 & 86.87 \\
iron & 38.9 & 99.80 & 32.03 & 85.02 & 57.9 & 98.88 & 97.85 & \textbf{100.0} \\
lamp & 35.1 & \textbf{99.71} & 60.47 & 57.26 & 54.2 & 99.33 & 97.89 & 96.84 \\
phone & 21.2 & 93.56 & 33.79 & 29.08 & 41.8 & 92.41 & 90.75 & \textbf{94.69} \\
\midrule
average & 30.2 & \textbf{96.48} & 28.65 & 50.00 & 49.5 & 86.27 & 89.86 & 95.15 \\
\bottomrule
\end{tabular}
\end{adjustbox}
\vspace{-4mm}
\end{center}
\caption{
  Comparison of our method to the baseline methods on the LINEMOD dataset.
  The accuracies are reported in terms of ADD(-S) metric. * indicates symmetric object.}
\vspace{-5mm}
\label{table:linemod_results}
\end{table*}

\vspace{1mm}
\noindent\textbf{Number of Annotated Images.}
The performance of our method heavily relies on how well we can align the 3D virtual keypoints with the object coordinate frame.
We test our method on different number of labeled images. Fig.~\ref{fig:boxplot} shows the results on ``cat" (left) and ``YCB-Can" (right).
``Train Samples" means we randomly sample the labeled images from training set (\ie seen during training), and ``Test Samples" means we randomly sample images from the original test dataset (\ie unseen during training).
For each number of pose 
labels, we run 10$\times$ and record the accuracies.

The variance of the accuracies from the trials is very small when the labeled images are included in the training set. This shows our method can reliably align the virtual keypoints with the object coordinate frame regardless of the number of labeled images or the performance of the trained pose estimator.
The performance highly depends on the accuracy of pose estimator when the labeled images are not used for training. For example, the variance is small on the more accurate ``Cat", and large on the ``YCB-Can".
The proposed keypoint alignment algorithm is able to reject outliers and achieve robust performance when there are enough number of labeled images.
Since our accuracy is not critically related to the number of pose labels when the network is trained on labeled images, we report the accuracy using all the training images ($\sim180$ on average) to align the virtual keypoints in the rest of the experiments.
Detailed results using few labeled images (\eg 1 and 3) are provided in the supplementary material.

\begin{table}[t]
\begin{center}
\small
\setlength{\tabcolsep}{4pt}
\begin{adjustbox}{max width=\columnwidth}
\begin{tabular}{ c | c  c  c  c | c}
\toprule
\multirow{2}{*}{Methods} & {PVNet} & {Pix2Pose} & {HybridPose} & {RLLG} & \multirow{2}{*}{\textbf{Ours}} \\
& \cite{peng2019pvnet} & \cite{park2019pix2pose} & \cite{song2020hybridpose} & \cite{cai2020reconstruct} &  \\
\midrule
ape & 15.81 & 22.0 & 20.9 & 7.1 & \textbf{25.63} \\
can & 63.30 & 44.7 & \textbf{75.3} & 40.6 & 65.76 \\
cat & 16.68 & 22.7 & 24.9 & 15.6 & \textbf{25.17} \\
driller & 65.65 & 44.7 & \textbf{70.2} & 43.9 & 67.12 \\
duck & 25.24 & 15.0 & \textbf{27.9} & 12.9 & 23.24 \\
eggbox* & 50.17 & 25.2 & \textbf{52.4} & 46.4 & 27.57 \\
glue* & 49.62 & 32.4 & \textbf{53.8} & 51.7 & 51.77 \\
holep &39.67 & 49.5 & \textbf{54.3} & 24.5 & 39.34 \\
\midrule
average & 40.77 & 32.0 & \textbf{47.5} & 30.3 & 40.70 \\
\bottomrule
\end{tabular}
\end{adjustbox}
\vspace{-3mm}
\end{center}
\caption{
  Comparison of our method to the baseline methods on the OCCLUSION dataset. The accuracies are reported in terms of ADD(-S) metric. * indicates symmetric object. All baseline methods are trained with ground truth poses.
}
\label{table:occlusion_results}
\vspace{-3mm}
\end{table}

\begin{figure*}[t]
\begin{center}
\includegraphics[width=\linewidth]{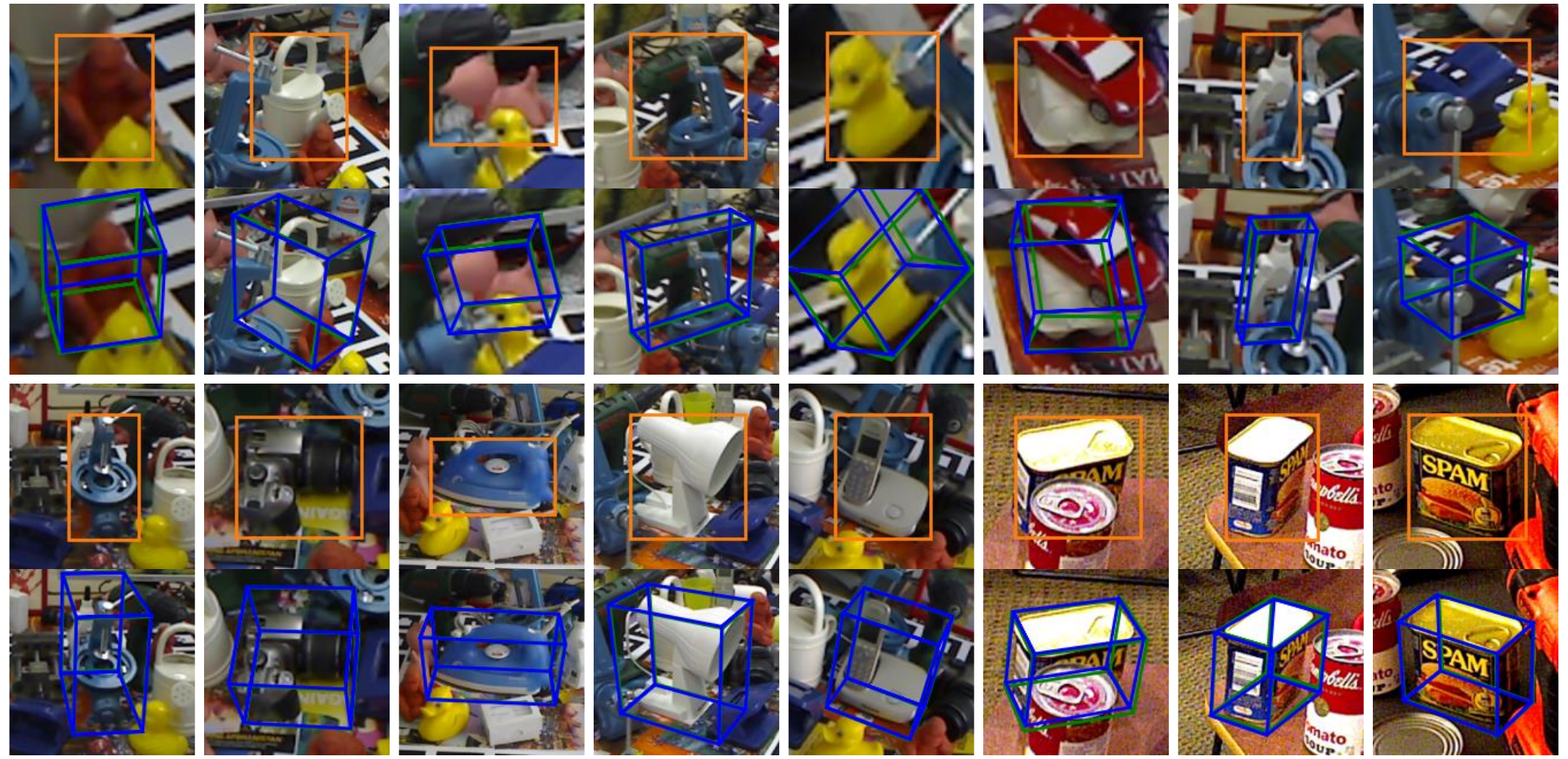}
\end{center}
\vspace{-5mm}
\caption{Visualization of the results on OCCLUSION (top row), LINEMOD and YCB-Can (bottom row). For each object, we show the RGB input to the keypoint detection network along with the orange box showing the 2D detection results. The ground truth 6D pose is shown in green, while our prediction is shown in blue.}
\label{fig:qual_results}
\vspace{-3mm}
\end{figure*}

\subsection{Comparison with Baselines}

\noindent \textbf{LINEMOD.}
In Tab.~\ref{table:linemod_results}, we compare our method with several state-of-the-art approaches trained on either synthetic or real data.
OK-POSE \cite{zhao2020learning} is the only previous work which shares exactly the same problem setting with us, \ie learning from relative transformations between viewpoints of image pairs.
Compared to this baseline, our method achieves significantly higher overall accuracy (+66.32\%).
OK-POSE formulates the problem as automatically discovering a set of distinct keypoints and lifting them to 3D space via depth regression.
The 6D object pose is inferred from the corresponding 3D keypoints in the query image and the reference image.
However, the task of learning keypoints that consistently locate on the object across different viewpoints is a difficult problem, especially when the object is texture-less and symmetric.
Depth regression is often shown to be less accurate than solving a PnP problem in the context of 6D object pose estimation.
Our formulation can avoid the above-mentioned problems and the result provides evidence for our advantage.

Our method even outperforms recent approaches (\eg PVNet \cite{peng2019pvnet}, CDPN \cite{li2019cdpn}, and DPOD \cite{zakharov2019dpod}) trained on real images annotated with full 6D object poses.
The marginal improvement is partially due to being a top-down approach, \ie locate the object on image first and then detect keypoints in the region of interest.
There exists a category of works, \eg AAE \cite{sundermeyer2018implicit}, DPOD \cite{zakharov2019dpod}, and NOL \cite{park2020neural}, which aims to reduce the labeling cost through training pose estimator on synthetic images rendered from 3D CAD model.
Theoretically, these methods can have unlimited amount of training data which covers every viewpoint.
However, the domain gap that hinders their performances remains as a challenge.
Compared to them, our approach utilizes easily obtainable labels but achieves significantly higher accuracy.

\vspace{1mm}
\noindent \textbf{OCCLUSION.}
Tab.~\ref{table:occlusion_results} summaries the comparison with PVNet \cite{peng2019pvnet}, Pix2Pose \cite{park2019pix2pose}, HybridPose \cite{song2020hybridpose}, and RLLG \cite{cai2020reconstruct} on the OCCLUSION dataset in terms of ADD(-S) metric.
Although these methods are trained with ground truth 6D pose labels, our method still achieves comparable performance.
The result also indicates that our method is less competitive when objects are occluded.
We argue that the lack of foreground mask information makes our keypoint detection network prone to foreground distractions.

\subsection{Qualitative Results}
Fig.~\ref{fig:qual_results} shows the qualitative results on OCCLUSION, LINEMOD and YCB-Can.
Our method is able to recover the 6D pose even when the objects are partially visible.

\section{Limitations}
Similar to \cite{zhao2020learning}, our method requires a reference image with 6D object pose for inference. Nonetheless, we only require one reference image for the entire inference stage (\cf Remark \ref{remark:oneRefImage}), which significantly outweighs the need of many labeled images for training. Furthermore, the 6D object pose of the one reference image can be easily obtained from epipolar geometry (\cf supplementary).

\section{Conclusions}
We present a novel weakly supervised keypoint-based approach for 6D object pose estimation.
Our keypoint detection network learns to detect the 2D projections of randomly generated 3D keypoints under the supervision of relative transformations between camera viewpoints.
During inference, the 3D keypoints are aligned with the object coordinate frame through a single reference image.
Extensive experiments demonstrate that our approach is comparable to state-of-the-art fully supervised approaches.

{\small
\bibliographystyle{ieee_fullname}
\bibliography{egbib}

\begin{thebibliography}{10}\itemsep=-1pt

\bibitem{bay2006surf}
Herbert Bay, Tinne Tuytelaars, and Luc Van~Gool.
\newblock Surf: Speeded up robust features.
\newblock In {\em European conference on computer vision}, pages 404--417.
  Springer, 2006.

\bibitem{bousmalis2017unsupervised}
Konstantinos Bousmalis, Nathan Silberman, David Dohan, Dumitru Erhan, and Dilip
  Krishnan.
\newblock Unsupervised pixel-level domain adaptation with generative
  adversarial networks.
\newblock In {\em Proceedings of the IEEE conference on computer vision and
  pattern recognition}, pages 3722--3731, 2017.

\bibitem{brachmann2014learning}
Eric Brachmann, Alexander Krull, Frank Michel, Stefan Gumhold, Jamie Shotton,
  and Carsten Rother.
\newblock Learning 6d object pose estimation using 3d object coordinates.
\newblock In {\em European conference on computer vision}, pages 536--551.
  Springer, 2014.

\bibitem{brachmann2016uncertainty}
Eric Brachmann, Frank Michel, Alexander Krull, Michael Ying~Yang, Stefan
  Gumhold, et~al.
\newblock Uncertainty-driven 6d pose estimation of objects and scenes from a
  single rgb image.
\newblock In {\em Proceedings of the IEEE conference on computer vision and
  pattern recognition}, pages 3364--3372, 2016.

\bibitem{cai2020reconstruct}
Ming Cai and Ian Reid.
\newblock Reconstruct locally, localize globally: A model free method for
  object pose estimation.
\newblock In {\em Proceedings of the IEEE/CVF Conference on Computer Vision and
  Pattern Recognition}, pages 3153--3163, 2020.

\bibitem{fischler1981random}
Martin~A Fischler and Robert~C Bolles.
\newblock Random sample consensus: a paradigm for model fitting with
  applications to image analysis and automated cartography.
\newblock {\em Communications of the ACM}, 24(6):381--395, 1981.

\bibitem{hartley2013rotation}
Richard Hartley, Jochen Trumpf, Yuchao Dai, and Hongdong Li.
\newblock Rotation averaging.
\newblock {\em International journal of computer vision}, 103(3):267--305,
  2013.

\bibitem{hartley2003multiple}
Richard Hartley and Andrew Zisserman.
\newblock {\em Multiple view geometry in computer vision}.
\newblock Cambridge university press, 2003.

\bibitem{he2016deep}
Kaiming He, Xiangyu Zhang, Shaoqing Ren, and Jian Sun.
\newblock Deep residual learning for image recognition.
\newblock In {\em Proceedings of the IEEE conference on computer vision and
  pattern recognition}, pages 770--778, 2016.

\bibitem{he2020pvn3d}
Yisheng He, Wei Sun, Haibin Huang, Jianran Liu, Haoqiang Fan, and Jian Sun.
\newblock Pvn3d: A deep point-wise 3d keypoints voting network for 6dof pose
  estimation.
\newblock In {\em Proceedings of the IEEE/CVF Conference on Computer Vision and
  Pattern Recognition}, pages 11632--11641, 2020.

\bibitem{hinterstoisser2012model}
Stefan Hinterstoisser, Vincent Lepetit, Slobodan Ilic, Stefan Holzer, Gary
  Bradski, Kurt Konolige, and Nassir Navab.
\newblock Model based training, detection and pose estimation of texture-less
  3d objects in heavily cluttered scenes.
\newblock In {\em Asian conference on computer vision}, pages 548--562.
  Springer, 2012.

\bibitem{hodavn2020bop}
Tom{\'a}{\v{s}} Hoda{\v{n}}, Martin Sundermeyer, Bertram Drost, Yann Labb{\'e},
  Eric Brachmann, Frank Michel, Carsten Rother, and Ji{\v{r}}{\'\i} Matas.
\newblock Bop challenge 2020 on 6d object localization.
\newblock In {\em European Conference on Computer Vision}, pages 577--594.
  Springer, 2020.

\bibitem{hu2018segmentation}
Yinlin Hu, Joachim Hugonot, Pascal Fua, and Mathieu Salzmann.
\newblock Segmentation-driven 6d object pose estimation.
\newblock In {\em Proceedings of the IEEE Conference on Computer Vision and
  Pattern Recognition}, pages 3385--3394, 2019.

\bibitem{iskakov2019learnable}
Karim Iskakov, Egor Burkov, Victor Lempitsky, and Yury Malkov.
\newblock Learnable triangulation of human pose.
\newblock In {\em Proceedings of the IEEE International Conference on Computer
  Vision}, pages 7718--7727, 2019.

\bibitem{glenn_jocher_2020_4154370}
Glenn Jocher.
\newblock ultralytics/yolov5: v3.1 - bug fixes and performance improvements.
\newblock \url{https://doi.org/10.5281/zenodo.4154370}, 2020.

\bibitem{kabsch1976solution}
Wolfgang Kabsch.
\newblock A solution for the best rotation to relate two sets of vectors.
\newblock {\em Acta Crystallographica Section A: Crystal Physics, Diffraction,
  Theoretical and General Crystallography}, 32(5):922--923, 1976.

\bibitem{kehl2017ssd}
Wadim Kehl, Fabian Manhardt, Federico Tombari, Slobodan Ilic, and Nassir Navab.
\newblock Ssd-6d: Making rgb-based 3d detection and 6d pose estimation great
  again.
\newblock In {\em Proceedings of the IEEE International Conference on Computer
  Vision}, pages 1521--1529, 2017.

\bibitem{kehl2016deep}
Wadim Kehl, Fausto Milletari, Federico Tombari, Slobodan Ilic, and Nassir
  Navab.
\newblock Deep learning of local rgb-d patches for 3d object detection and 6d
  pose estimation.
\newblock In {\em European Conference on Computer Vision}, pages 205--220.
  Springer, 2016.

\bibitem{li2018unified}
Chi Li, Jin Bai, and Gregory~D Hager.
\newblock A unified framework for multi-view multi-class object pose
  estimation.
\newblock In {\em Proceedings of the European Conference on Computer Vision
  (ECCV)}, pages 254--269, 2018.

\bibitem{li2019cdpn}
Zhigang Li, Gu Wang, and Xiangyang Ji.
\newblock Cdpn: Coordinates-based disentangled pose network for real-time
  rgb-based 6-dof object pose estimation.
\newblock In {\em Proceedings of the IEEE International Conference on Computer
  Vision}, pages 7678--7687, 2019.

\bibitem{lowe1999object}
David~G Lowe.
\newblock Object recognition from local scale-invariant features.
\newblock In {\em Proceedings of the seventh IEEE international conference on
  computer vision}, volume~2, pages 1150--1157. Ieee, 1999.

\bibitem{manhardt2019explaining}
Fabian Manhardt, Diego~Martin Arroyo, Christian Rupprecht, Benjamin Busam,
  Tolga Birdal, Nassir Navab, and Federico Tombari.
\newblock Explaining the ambiguity of object detection and 6d pose from visual
  data.
\newblock In {\em Proceedings of the IEEE International Conference on Computer
  Vision}, pages 6841--6850, 2019.

\bibitem{oberweger2018making}
Markus Oberweger, Mahdi Rad, and Vincent Lepetit.
\newblock Making deep heatmaps robust to partial occlusions for 3d object pose
  estimation.
\newblock In {\em Proceedings of the European Conference on Computer Vision
  (ECCV)}, pages 119--134, 2018.

\bibitem{park2019pix2pose}
Kiru Park, Timothy Patten, and Markus Vincze.
\newblock Pix2pose: Pixel-wise coordinate regression of objects for 6d pose
  estimation.
\newblock In {\em Proceedings of the IEEE International Conference on Computer
  Vision}, pages 7668--7677, 2019.

\bibitem{park2020neural}
Kiru Park, Timothy Patten, and Markus Vincze.
\newblock Neural object learning for 6d pose estimation using a few cluttered
  images.
\newblock In {\em Proceedings of the European Conference on Computer Vision
  (ECCV)}, 2020.

\bibitem{peng2019pvnet}
Sida Peng, Yuan Liu, Qixing Huang, Xiaowei Zhou, and Hujun Bao.
\newblock Pvnet: Pixel-wise voting network for 6dof pose estimation.
\newblock In {\em Proceedings of the IEEE Conference on Computer Vision and
  Pattern Recognition}, pages 4561--4570, 2019.

\bibitem{planche2019seeing}
Benjamin Planche, Sergey Zakharov, Ziyan Wu, Andreas Hutter, Harald Kosch, and
  Slobodan Ilic.
\newblock Seeing beyond appearance-mapping real images into geometrical domains
  for unsupervised cad-based recognition.
\newblock In {\em 2019 IEEE/RSJ International Conference on Intelligent Robots
  and Systems (IROS)}, pages 2579--2586. IEEE, 2019.

\bibitem{rad2017bb8}
Mahdi Rad and Vincent Lepetit.
\newblock Bb8: A scalable, accurate, robust to partial occlusion method for
  predicting the 3d poses of challenging objects without using depth.
\newblock In {\em Proceedings of the IEEE International Conference on Computer
  Vision}, pages 3828--3836, 2017.

\bibitem{rad2018domain}
Mahdi Rad, Markus Oberweger, and Vincent Lepetit.
\newblock Domain transfer for 3d pose estimation from color images without
  manual annotations.
\newblock In {\em Asian Conference on Computer Vision}, pages 69--84. Springer,
  2018.

\bibitem{rad2018feature}
Mahdi Rad, Markus Oberweger, and Vincent Lepetit.
\newblock Feature mapping for learning fast and accurate 3d pose inference from
  synthetic images.
\newblock In {\em Proceedings of the IEEE Conference on Computer Vision and
  Pattern Recognition}, pages 4663--4672, 2018.

\bibitem{redmon2018yolov3}
Joseph Redmon and Ali Farhadi.
\newblock Yolov3: An incremental improvement.
\newblock {\em arXiv preprint arXiv:1804.02767}, 2018.

\bibitem{song2020hybridpose}
Chen Song, Jiaru Song, and Qixing Huang.
\newblock Hybridpose: 6d object pose estimation under hybrid representations.
\newblock In {\em Proceedings of the IEEE/CVF Conference on Computer Vision and
  Pattern Recognition}, pages 431--440, 2020.

\bibitem{sun2018integral}
Xiao Sun, Bin Xiao, Fangyin Wei, Shuang Liang, and Yichen Wei.
\newblock Integral human pose regression.
\newblock In {\em Proceedings of the European Conference on Computer Vision
  (ECCV)}, pages 529--545, 2018.

\bibitem{sundermeyer2018implicit}
Martin Sundermeyer, Zoltan-Csaba Marton, Maximilian Durner, Manuel Brucker, and
  Rudolph Triebel.
\newblock Implicit 3d orientation learning for 6d object detection from rgb
  images.
\newblock In {\em Proceedings of the European Conference on Computer Vision
  (ECCV)}, pages 699--715, 2018.

\bibitem{tekin2018real}
Bugra Tekin, Sudipta~N Sinha, and Pascal Fua.
\newblock Real-time seamless single shot 6d object pose prediction.
\newblock In {\em Proceedings of the IEEE Conference on Computer Vision and
  Pattern Recognition}, pages 292--301, 2018.

\bibitem{wang2019densefusion}
Chen Wang, Danfei Xu, Yuke Zhu, Roberto Mart{\'\i}n-Mart{\'\i}n, Cewu Lu, Li
  Fei-Fei, and Silvio Savarese.
\newblock Densefusion: 6d object pose estimation by iterative dense fusion.
\newblock In {\em Proceedings of the IEEE Conference on Computer Vision and
  Pattern Recognition}, pages 3343--3352, 2019.

\bibitem{wang2020self6d}
Gu Wang, Fabian Manhardt, Jianzhun Shao, Xiangyang Ji, Nassir Navab, and
  Federico Tombari.
\newblock Self6d: Self-supervised monocular 6d object pose estimation.
\newblock In {\em Proceedings of the European Conference on Computer Vision
  (ECCV)}, 2020.

\bibitem{xiang2018posecnn}
Yu Xiang, Tanner Schmidt, Venkatraman Narayanan, and Dieter Fox.
\newblock Posecnn: A convolutional neural network for 6d object pose estimation
  in cluttered scenes.
\newblock In {\em Robotics: Science and Systems (RSS)}, 2018.

\bibitem{xiao2010sun}
Jianxiong Xiao, James Hays, Krista~A Ehinger, Aude Oliva, and Antonio Torralba.
\newblock Sun database: Large-scale scene recognition from abbey to zoo.
\newblock In {\em 2010 IEEE computer society conference on computer vision and
  pattern recognition}, pages 3485--3492. IEEE, 2010.

\bibitem{zakharov2019deceptionnet}
Sergey Zakharov, Wadim Kehl, and Slobodan Ilic.
\newblock Deceptionnet: Network-driven domain randomization.
\newblock In {\em Proceedings of the IEEE International Conference on Computer
  Vision}, pages 532--541, 2019.

\bibitem{zakharov2019dpod}
Sergey Zakharov, Ivan Shugurov, and Slobodan Ilic.
\newblock Dpod: 6d pose object detector and refiner.
\newblock In {\em Proceedings of the IEEE International Conference on Computer
  Vision}, pages 1941--1950, 2019.

\bibitem{zhao2020learning}
Wanqing Zhao, Shaobo Zhang, Ziyu Guan, Wei Zhao, Jinye Peng, and Jianping Fan.
\newblock Learning deep network for detecting 3d object keypoints and 6d poses.
\newblock In {\em Proceedings of the IEEE/CVF Conference on Computer Vision and
  Pattern Recognition}, pages 14134--14142, 2020.

\end{thebibliography}
}

\renewcommand\thesection{\Alph{section}}
\setcounter{section}{0}
\section{Justifications for Our Problem Setting}
\paragraph{Training: Getting the relative transformations.} In contrast to the fully supervised setting which requires ground truth 6D object poses and the fine-grained 3D object model during training, our weakly supervised method is trained only on image pairs with known relative transformations between camera viewpoints.
Compared to the ground truth 6D pose, the relative transformation between the viewpoints of an image pair is easier to obtain.
For example, the relative transformation can be obtained from the camera calibration for an image pair captured by a binocular camera.
The relative transformation can also be measured by the inertial navigation system (INS) equipped on most smart phones or tablets.
In the scenario of robotic manipulation, the camera is often mounted on a robotic arm. The relative transformation can also be computed from forward kinematics.

\vspace{-4mm}
\paragraph{Inference: Getting the reference image with 6D object pose.} Similar to the inference setting of OK-Pose \cite{zhao2020learning}, our method needs a reference image with ground truth 6D pose to retrieve the absolute pose information.
This ground truth 6D pose of the reference image can be obtained easily \textit{without} reconstructing the full 3D model of the object.
For example, we can select the reference image from a training image pair. Given the pair of images, we manually label the 2D keypoint correspondences ($\geq 4$ for a unique solution from PnP later) of the object. The corresponding 3D points are then computed using triangulation with the known relative pose between the image pairs. Once a proper coordinate frame (depends on user or application) is assigned to the 3D points, the ground truth 6D pose can be obtained via solving the PnP problem. 
Although this method can also be used to get the 6D object pose and 3D keypoints for fully supervised learning, the manual labeling of the 2D keypoint correspondences is laborious and costly for large amounts of training data needed for full supervision. In contrast, we only need one reference image for the entire inference stage. 

\vspace{-4mm}
\paragraph{Why not use image keypoints such as SIFT \cite{lowe1999object} or SURF \cite{bay2006surf}?} At first sight, the obvious question is why not we simply replace the laborious 2D keypoint labeling with keypoint detectors and descriptors such as SIFT or SURF? In this way, we can easily generate large amounts of ground truth 6D object poses and 3D keypoints for fully supervised training. However, there are two limitations of this approach: 1) It is not easy to determine which SIFT or SURF 2D keypoints belong to the 3D object. An object detector, \eg YOLOv3 \cite{redmon2018yolov3} can help to mitigate this problem, but it is still difficult to guarantee high accuracy (we require almost 100\% accuracy as this is for the ground truth labels) 
in a cluttered scene. 2) 2D keypoints detector and descriptor are not sufficiently robust to ensure fully accurate detection of the correspondences. The results can be highly susceptible to false positive correspondences, especially in scenes with a lot of repetitive appearances. Consequently, laborious quality checks are still needed to ensure the assigned labels are of high accuracy for fully supervised learning.          

\vspace{-4mm}
\paragraph{No need for 3D model of the objects.}
It is interesting to note that there is no need for our problem setting to be given nor reconstruct any 3D model of the objects. For training, the ground truths relative transformations can be obtained directly by binocular camera calibration, INS system, forward kinematics, etc. For inference, the 6D absolute pose of the one reference image can be easily obtained via epipolar geometry and PnP with at least 4 manually labeled 2D image keypoint correspondences over a pair of images with known relative transformation, \eg an image pair from the training dataset. In this process, we just have to triangulate for the $\geq 4$ 3D points on the object. There is no need to reconstruct the full 3D model of the object.

\section{Average Offset Transformation}
Given a set of offset transformations $\{(R^i_a, \mathbf{t}^i_a)\}$ computed from the reference images, Algorithm~\ref{alg:averaging} computes the average offset rotation $\bar{R}_a$ and translation $\bar{\mathbf{t}}_a$ by taking the geodesic $L_2\text{-Mean}$ and the arithmetic mean, respectively.

\begin{algorithm}[ht]
\DontPrintSemicolon
\SetKwInOut{Input}{Input}
\SetKwInOut{Output}{Output}
\SetKwInOut{Require}{Require}
\Require{$t_r$, $t_t$, $\eta$, \textit{max\_iter}, $\epsilon$, \textit{max\_step}}
\Input{$\{(R^i_a, \mathbf{t}^i_a)\}, i \in \{1, 2, \dots, N \}$}
\Output{$\bar{R}_a \in SO(3), \bar{\mathbf{t}}_a \in \mathbb{R}^3$}
initialize inlier set $\mathbb{S} := \{\}$, inlier ratio $r := 0.0$ \\
\For{$\text{iter} := 1$ \KwTo max\_iter}{
  randomly select $k$ from $\{1, 2, \dots N \}$\\
  $\mathbb{I} := \{\}$ \\
  \For{$i := 1$ \KwTo $N$}{
    $\Delta r := \| \log(R^k_a{^T} R^i_a) \|_2$ \\
    $\Delta t := \| \mathbf{t}^k_a - \mathbf{t}^i_a \|_2$ \\
    \lIf{$\Delta r < t_r \land \Delta t < t_t$} {$\mathbb{I} := \{i\} \cup \mathbb{I}$}
  }
  \If{$| \mathbb{S} | < | \mathbb{I} |$}{
    $\mathbb{S} := \mathbb{I}$ \\
    $r := |\mathbb{S}| / N$
  }
  \lIf{$(1 - (1 - r) ^ {\text{iter}}) > \eta$} {break}
}
$\bar{R}_a := R^j_a$, $\forall j \in \mathbb{S}$ \\
\For{$\text{step} := 1$ \KwTo max\_step}{
  $\mathbf{r} := \frac{1}{|\mathbb{S}|}\sum_{j \in \mathbb{S}} \log(\bar{R}^T_a R^j_a)$ \\
  \lIf{$\| \mathbf{r} \|_2 < \epsilon$} {break}
  $\bar{R}_a := \bar{R}_a \exp(\mathbf{r})$ \\
}
$\bar{\mathbf{t}}_a := \frac{1}{|\mathbb{S}|} \sum_{j \in \mathbb{S}} \mathbf{t}^j_a$ \\
\caption{Compute average transformation.}
\label{alg:averaging}
\end{algorithm}

\section{Additional Results}

In Tab.~\ref{table:lm_results} and \ref{table:occ_results}, we provide the results on LINEMOD and OCCLUSION when using only 1 or 3 labeled images to compute the offset transformation $T_a$.
For each object, we run the evaluation 10 times (re-sample labeled images each time) and report the average accuracy.
It is shown that there is not much difference between using all training images and using single labeled image.

We also report the accuracies in terms of 2D projection metric and $n ^\circ \, m \text{cm}$ metric.
The 2D projection metric computes the mean distance between the projections of 3D model points under the predicted pose and the ground truth pose.
A pose is considered as correct if the mean distance is less than 5 pixels.
The $n ^\circ \, m \text{cm}$ metric directly measures the rotational and transnational errors.
We set the thresholds for a correct pose to $5 ^\circ \, 5 \text{cm}$.

\begin{table}[t]
\begin{center}
\setlength{\tabcolsep}{4pt}
\begin{adjustbox}{max width=\columnwidth}
\begin{tabular}{ c | c  c  c  c  c }
\toprule
metric & ADD-1 & ADD-3 & ADD-all & Proj-5 & $5^\circ 5 \text{cm}$ \\
\midrule
ape & 88.54 & 88.95 & 89.14 & 98.95 & 98.95  \\
bvise & 99.61 & 99.61 & 99.61 & 99.71 & 98.55  \\
cam & 97.96 & 98.04 & 98.14 & 98.82 & 99.31  \\
can & 99.00 & 99.01 & 99.02 & 99.70 & 99.31 \\
cat & 96.08 & 96.79 & 97.01 & 99.50 & 99.80  \\
driller & 99.07 & 99.04 & 99.01 & 98.61 & 99.01  \\
duck & 86.32 & 86.92 & 87.14 & 98.22 & 97.93  \\
eggbox* & 100.0 & 100.0 & 100.0 & 99.44 & 100.0  \\
glue* & 99.37 & 99.41 & 99.42 & 99.61 & 98.07  \\
holp & 92.57 & 92.85 & 92.67 & 99.81 & 97.34  \\
iron & 99.86 & 99.81 & 99.80 & 99.39 & 98.26  \\
lamp & 99.68 & 99.67 & 99.71 & 97.79 & 98.94 \\
phone & 93.03 & 93.50 & 93.56 & 99.23 & 97.79  \\
\midrule
average & 96.24 & 96.43 & 96.48 & 99.14 & 98.71  \\
\bottomrule
\end{tabular}
\end{adjustbox}
\end{center}
\vspace{-5mm}
\caption{Accuracies on the LINEMOD dataset in terms of different evaluation metrics. * indicates symmetric object.} \vspace{-5mm}
\label{table:lm_results}
\end{table}

\begin{table}[t]
\begin{center}
\setlength{\tabcolsep}{4pt}
\begin{adjustbox}{max width=\columnwidth}
\begin{tabular}{ c | c  c  c  c  c }
\toprule
metric & ADD-1 & ADD-3 & ADD-all & Proj-5 & $5^\circ 5 \text{cm}$ \\
\midrule
ape & 25.13 & 25.88 & 25.63 & 62.36 & 40.97  \\
can & 65.80 & 65.75 & 65.76 & 77.37 & 37.95  \\
cat & 24.70 & 25.00 & 25.17 & 65.21 & 20.12  \\
driller & 66.81 & 66.98 & 67.12 & 70.81 & 46.94 \\
duck & 22.52 & 23.08 & 23.24 & 56.61 & 13.22  \\
eggbox* & 27.63 & 27.69 & 27.57 & 3.22 & 0.20  \\
glue* & 51.22 & 51.62 & 51.77 & 55.85 & 16.43  \\
holep & 39.89 & 39.06 & 39.34 & 67.77 & 24.05  \\
\midrule
average & 40.46 & 40.63 & 40.70 & 57.40 & 24.99  \\
\bottomrule
\end{tabular}
\end{adjustbox}
\end{center}\vspace{-5mm}
\caption{Accuracies on the OCCLUSION dataset in terms of different evaluation metrics. * indicates symmetric object.} \vspace{-5mm}
\label{table:occ_results}
\end{table}

\section{Runtime Analysis}
Given a $480 \times 640$ image, our approach runs at 27fps on a desktop with an Intel Core i7-5960X CPU (3.0GHz) and a NVIDIA GTX 1080Ti GPU.
Specifically, it takes about 30ms for 2D object detection, 5.8ms for keypoints detection, and 0.2ms for solving PnP.
Considering that object detection is the bottleneck of our pipeline, the inference speed can be compensated by utilizing a lightweight object detector (e.g. tiny-YOLOv3).

\section{Failure Cases}
There are two major sources of failure cases: missing 2D object detection and severe occlusion.
The trained 2D object detector achieves an accuracy of $99.9\%$ on LINEMOD, and an accuracy of $89.8\%$ on OCCLUSION ($\text{IoU} > 0.5$).
If an object is missed by the 2D object detector, there is no chance to correctly recover its 6D pose.
Our problem setting does not provide mask labels for objects. It is thus difficult to make it robust to severe occlusions since current robust approaches are based on dense predictions which require some sort of foreground information.

\end{document}